\documentclass[conference,10pt]{IEEEtran}
\IEEEoverridecommandlockouts
\usepackage{xcolor}
\usepackage{tikz}
\usepackage{pgfplots}
\usepackage[ruled]{algorithm2e}
\usepackage{amsmath,epsfig,amssymb,amsthm,cite,url,xcolor}
\usepackage{hyperref}
\usepackage{enumerate}
\usepackage[latin1]{inputenc}

\theoremstyle{plain}
\newtheorem{theorem}{Theorem}
\newtheorem{assumption}{Assumption}
\newtheorem{lemma}[theorem]{Lemma}

\theoremstyle{definition}

\newtheoremstyle{specialcasestyle}{1mm}{1mm}{\upshape}{}{\bfseries\upshape}{.}{0mm}{}
\theoremstyle{specialcasestyle}

\newcommand{\figref}[1]{Fig.~\protect\ref{#1}}

\newcommand{\bu}{{\bf u}}

\newcommand{\bx}{{\bf x}}
\newcommand{\bv}{{\bf v}}

\newcommand{\by}{{\bf y}}
\newcommand{\bw}{{\bf w}}

\newcommand{\bz}{{\bf z}}

\newcommand{\bI}{{\bf I}}

\newcommand{\bm}{{\bf m}}

  \newcommand{\gm}{{\rm geomed}}

     \makeatletter
\def\endthebibliography{%
  \def\@noitemerr{\@latex@warning{Empty `thebibliography' environment}}%
  \endlist
}
\makeatother
\usepackage{graphicx}
\usepackage{placeins}
\usepackage{float}
\usepackage{tabularx}
\usepackage{tkz-euclide,subfigure}
\usepackage{amsmath,epsfig,amssymb}
\usepackage{amsmath}
\usepackage{bbm}
\usetikzlibrary{intersections}
\usepgfplotslibrary{fillbetween}

\pgfplotsset{compat=1.16} 
\begin{document}

\title{Robust Federated Learning via Over-The-Air Computation
}

\author{\IEEEauthorblockN{Houssem Sifaou and Geoffrey Ye Li}

\IEEEauthorblockA{{Department of Electrical and Electronic Engineering,}
{Imperial College London, UK}\\
\textit{emails:  \{h.sifaou,geoffrey.li\}@imperial.ac.uk}
}}

\maketitle

\begin{abstract}
This paper investigates the robustness of over-the-air federated learning to Byzantine attacks. The simple averaging of the model updates via over-the-air computation makes the learning task vulnerable to random or intended modifications of the local model updates of some malicious clients. We propose a robust transmission and aggregation framework to such attacks while preserving the benefits of over-the-air computation for federated learning. For the proposed robust federated learning, the participating clients are randomly divided into groups and a transmission time slot is allocated to each group. The parameter server aggregates the results of the different groups using a robust aggregation technique and conveys the result to the clients for another training round. We also analyze the convergence of the proposed algorithm. Numerical simulations confirm the robustness of the proposed approach to Byzantine attacks. 
\end{abstract}

\section{Introduction}

With the rapid growth and unprecedented success of artificial intelligence (AI) applications, a huge amount of data is transferred every day from distributed clients to data centers for further processing. However, in many applications, the data is sensitive, and sending it to data centers constitutes a major privacy concern \cite{10.1145/335191.335438,duchi2013local,Zhou2018}. One promising solution is federated learning (FL)\cite{pmlr-v54-mcmahan17a,DBLP:journals/corr/abs-1912-04977,9530714}. With FL, multiple distributed clients train a global machine learning model without sharing their local data. Local computations are carried out at the different clients, and only the model parameters updates are sent to the central server (CS). The latter aggregates the local updates and forwards the result to the distributed clients for another training round.

Although data privacy is preserved using FL, allowing the clients to perform model parameter update opens the way for possible adversarial attacks. Some of the distributed clients may send modified parameter updates with the intention of misleading the learning process \cite{6582732,8454396}. In this context, Byzantine attacks are a popular class, where certain clients aim to prevent the model from converging or causing convergence to a falsified model. The Byzantine clients may act independently or collectively. Unfortunately, even a single malicious client in a distributed setup as in FL can seriously affect the end model \cite{blanchard2017machine}. Developing countermeasures to these attacks has gained an increasing interest in recent years. Considering the FL setup without communication constraint, several aggregation techniques have been proposed to robustify the stochastic gradient descent (SGD) in the distributed setup. Stochastic algorithms tolerating few Byzantine attacks has been developed by aggregating the stochastic gradients updates using the median \cite{Xie2018GeneralizedBS}, geometric median\cite{minsker2015geometric,10.1145/3154503}, trimmed mean \cite{yin2018byzantine}, and iterative filtering \cite{10.1145/3322205.3311083}. The Krum aggregation in \cite{blanchard2017machine} selects the stochastic gradient having a minimum distance from a certain number of nearest stochastic gradients. A robust stochastic aggregation (RSA) has been developed in \cite{li2019rsa}, which tolerates heterogeneous datasets and Byzantine attacks. Other related works include leveraging redundancy of gradients to enhance robustness \cite{chen2018draco,Rajput2019DETOXAR} and avoiding saddle points in non-convex distributed learning in presence of  Byzantine attacks \cite{yin2019defending}. Moreover, the advantages of reducing the variance of the stochastic gradients to defend against Byzantine attackers have been investigated in \cite{9153949}. All these works are applicable when the individual local updates are sent separately to the parameter server. However, in the case of over-the-air FL (OTA-FL) \cite{Kai2020,9042352}, the local model updates are sent simultaneously over the analog wireless channel. This makes the aforementioned robust aggregation techniques not directly applicable in the case of OTA-FL.

In this work, we investigate the problem of Byzantine attacks in the case of OTA-FL. Particularly, we propose a transmission and aggregation approach that simultaneously exploits the benefits of over-the-air computation while being robust to Byzantine attacks. By dividing the participating clients into several groups at each global training round, assigning a transmission time slot for each group, and aggregating the model updates of the different groups using geometric median aggregation. Theoretical convergence analysis of the proposed approach is conducted under some assumptions on the loss functions. The analysis reveals that when the number of attacks is less than half of the number of groups, the proposed algorithm converges at a linear rate to a neighborhood of the optimal solution with an asymptotic learning error that depends on the noise variance and the number of Byzantine attackers. Moreover, as evidenced by numerical results, the proposed approach is robust to Byzantine attacks compared with simple averaging OTA-FL. 

The remainder of the paper is organized as follows. In the next section, we introduce OTA-FL and the transmission model. In Section \ref{Proposed_approach}, our proposed robust approach is presented, while in Section \ref{conver_analysis} the convergence of the proposed algorithm is studied. Numerical results are provided in Section \ref{numerical_simulation} and concluding remarks are drawn in Section \ref{conclusion}.

\section{System Model}
\label{system_model}
We consider a FL system composed of a CS and $N$ clients. Client $n$ has its local dataset $D_n = \{(\bx_i\in \mathbb{R}^d ,y_i \in\mathbb{R})\}_{i=1}^{m_n}$ composed of $m_n$ samples. All the clients collaboratively train a global model by communicating with the CS. More precisely, they seek for the optimal parameter vector $\bw^\star\in\mathbb{R}^p$ that minimizes a global loss function $f(\bw)$ given by
\begin{equation}
f(\bw)=\frac{1}{N}\sum_{n=1}^N f_n(\bw),
\end{equation}
where $f_n(\bw)$ is the local loss function at client $n$ defined as
\begin{equation}
f_n(\bw)=\frac{1}{m_n}\sum_{j=1}^{m_n} \ell(\bw; \bx_j,y_j).
\end{equation}
Usually, the CS communicates with the clients using wireless channels. We consider in this work the case of analog OTA-FL \cite{9042352}, which will be introduced hereafter. 
\subsection{Analog over-the-air FL}
At each global training round $t$, the CS sends the model parameter vector, $\bw_t$, to the clients. It is usually assumed that the downlink communication is perfect due to the high power available at the CS. Thus, each client receives the global model without distortions. Then, client $n$  sets its local model as $\bw_{t,0}^n=\bw_t$ and runs its local SGD for $H$ iterations based on its local dataset 
\begin{equation}
\bw_{t,i+1}^n=\bw_{t,i}^n-\eta_t  f_{n,i_n^t}'(\bw_{t,i}^n), \ \ {\rm for} \ \ i = 0,1,\cdots, H-1,
\label{SGD}
\end{equation}
where $\eta_t$ is the SGD step size at round $t$ and $f_{n,i_n^t}'(\bw_{t,i}^n)$ denotes the stochastic gradient computed using a sample with index $i_n^t$ chosen uniformly at random from the local dataset of client $n$. In practice, 
After $H$ iterations, the clients convey their model updates given by
\begin{equation}
\bm_t^n = \bw_{t,H}^n - \bw_{t},
\end{equation}
simultaneously to the CS via analog OTA-FL. The model updates should be precoded in order to mitigate the effect of channel fading.  Let $ \tilde h_{n,t}= h_{t,n}e^{j \phi_t^n}$ be the a block fading channel corresponding to user $n$ at the transmission time of global round $t$, where $h_{t,n}  >0$ and $\phi_t^n=[-\pi,\pi]$ are its module and phase respectively. As in \cite{Kai2020,9076343,9042352,Sery2020,Liu2021}, we assume that perfect channel state information (CSI) is available at the clients and the CS. The imperfect CSI case is left for future investigation. Moreover, since the power budget at the clients is limited, the transmitted signal should satisfy the following average power constraint
\begin{equation}
\mathbb{E} \left[ \| \bx_n \|^2 \right] \leq P.
\label{power_const}
\end{equation}

In practice, weak channels might cause a high amplification of transmit power, possibly violating the transmission power constraint \eqref{power_const}. To overcome this issue, a threshold $h_{\min}$ can be set, and clients with channel fading coefficients less than $h_{\min}$ in magnitude will not transmit in that training round.

We adopt in this paper the precoding scheme proposed in \cite{Sery2020}, where every client $n$ precodes its model update $\bm_t^n$ as
\begin{equation}
 \bx_{t,n}=\begin{cases}\rho_{t}\frac{h_{ \rm min}}{ h_{t,n}}   e^{-j \phi_t^n}   \bm_t^n, \ \ \  {\rm if} \ \  h_{t,n}>h_{\min}\\ 0, \ \ \ \ \ \ \ \ \ \ \ \ \  \  \ \ \ \ \ \ \ {\rm if} \ \  h_{t,n} \leq h_{\min}\end{cases}
 \label{precoding}
\end{equation}
where $\rho_{t}$ is a factor to satisfy the average power constraint at client $n$. As proposed in \cite{Sery2020}, $\rho_t$ can be set as follows 
\begin{equation}
\rho_{t} =\sqrt{ \frac{P}{ \max_{n} \mathbb{E}  \| \bm_t^n \|^2}}.
\end{equation}
Note that in practice, the clients do not have access the updates of each  other in order to compute $\rho_t$ at each training round. A possible way to deal with this issue is that  the CS can estimate this parameter offline using a small dataset and then forward it to the clients so they can use it at every training round. Another solution is to obtain an upper bound for $\rho_t$ and use it at every iteration \cite{Sery2020}.
The received signal at the CS is 
\begin{equation}
\by_t=\sum_{n \in \mathcal{K}_t} \rho_t h_{\min}\bm^n_{t} +\tilde\bz_t,
\end{equation}
where $\mathcal{K}_t$ is the set of clients indices with channel fading verifying $h_{n,t}>h_{\min}$ and $\tilde\bz_t \sim \mathcal{N}(\boldsymbol{0},\sigma^2\bI_p)$ stands for additive noise. In order to update the global model, the CS sets
\begin{equation}
\bw_{t+1}=\frac{\by_t}{|\mathcal{K}_t|\rho_t h_{\min}}+\bw_t,
\label{global_update}
\end{equation}
where $|\mathcal{K}_t|$ is the cardinality of the set $\mathcal{K}_t$. The global model update in \eqref{global_update} can be also written as
\begin{equation}
\bw_{t+1}=\frac{1}{|\mathcal{K}_t|}\sum_{n=1}^{|\mathcal{K}_t| }\bw^n_{t,H} + \bz_t,
\end{equation}
where $\bz_t \triangleq  \frac{\tilde\bz_t}{|\mathcal{K}_t|\rho_t h_{\min}} \sim \mathcal{N}(\boldsymbol{0}, \frac{\sigma^2}{ |\mathcal{K}_t|^2 \rho_t^2 h_{\min}^2 } \bI_p)$.
\subsection{Byzantine attacks}

We assume that $B<N$ clients are malicious; sending arbitrary or modified parameter vector updates aiming at affecting the convergence of the global model or forcing it to converge to some particular solution. This type of attacks is known as Byzantine attacks. There are many works that proposed solutions to deal with this type of attacks in federated learning \cite{blanchard2017machine,Xie2018GeneralizedBS,minsker2015geometric,10.1145/3154503,yin2018byzantine,10.1145/3322205.3311083,li2019rsa,chen2018draco,Rajput2019DETOXAR,yin2019defending,9153949}. However, these works considered the case of wired FL or the case that each individual update of the client is sent separately to the CS. In this work, we consider the effect of Byzantine attacks in the context of analog OTA-FL and we propose a practical approach to deal with such attacks.

\section{Proposed approach}
\label{Proposed_approach}
In order to reduce the effect of the malicious model updates, we propose the following approach. At each global training round $t$, the CS divides uniformly at random the $N$ clients into $G=N/m$ groups where each group is composed of $m$ clients. Each group will be allocated a time slot for transmission of their model updates. Precisely, the clients of group $g$ will transmit simultaneously their updates over-the-air. This allows the CS to obtain $G$ model updates. Then, with a robust aggregation technique, the different model updates of the groups will be aggregated to update the global model. This approach will be robust to Byzantine attacks.

At global iteration $t$, the global model, $\bw_t$, is forwarded to all the clients. The clients in group $g$ perform $H_g$ steps of SGD using their local datasets as in $\eqref{SGD}$, and compute their model updates
\begin{equation}
\bm_t^n = \bw_{t,H_g}^n - \bw_{t}, \ \ {\rm for} \ \  n \in \mathcal{G}_{g,t},
\end{equation}
where $\mathcal{G}_{g,t}$ is the set of user indices belonging to group $g$ at the global training round $t$. Note that since the clients in different groups are not sending at the same time slot, we can let the number of the SGD steps varying among groups. In other terms, the clients of a group with later transmission time can perform more SGD steps than those in the current transmitting group. However, for simplicity we assume in the sequel that all clients perform the same number of SGD steps $H$ regardless of their transmission time.  
The clients in group $g$ compute their precoded signal as 
\begin{equation}
 \bx_{t,n}=\begin{cases}\rho_{t}\frac{h_{ \rm min}}{ h_{t,n}}   e^{-j \phi_t^n}   \bm_t^n, \ \ \  {\rm if} \ \  h_{t,n}>h_{\min}\\ 0, \ \ \ \ \ \ \ \ \ \ \ \ \  \   \ \ \ \ \ \ \ {\rm if} \ \  h_{t,n} \leq h_{\min}\end{cases}
 \label{precoding_11}
 \end{equation}
and send their updates simultaneously during their allocated transmission time slot. At the CS, the received signal vector corresponding to group $g$ can be expressed as
\begin{equation}
\by_{t,g}=\sum_{n \in \mathcal{K}_{t,g}} \rho_t h_{\min}\bm^n_{t} +\tilde\bz_{t,g},
\end{equation}
where $\mathcal{K}_{t,g}$ is the set of clients indices of group $g$ with channels such that $h_{n,t}>h_{\min}$ and $\tilde\bz_{t,g} \sim \mathcal{N}(\boldsymbol{0},\sigma^2\bI_p)$ is the additive noise. The CS estimates the model update of the group $g$ as
\begin{equation}
\bu_g ^t= \frac{\by_{t,g}}{\rho_t h_{\min} |\mathcal{K}_{t,g}|}.
\label{group_update}
\end{equation}
After all the group updates are collected, the CS disposes of $G$ vector updates $\bu_1^t,\cdots,\bu_G^t$ and can aggregate these updates to obtain the global model. For instance, one of the most efficient aggregation techniques that can be used is the geometric median \cite{bhagoji2019analyzing}. Other aggregation techniques can be used such as the Krum aggregation rule proposed in \cite{blanchard2017machine}. In this work, we focus on geometric median aggregation.
The global model is updated as
\begin{equation}
\bw_{t+1} = \gm(\bu_1^t,\cdots,\bu_G^t) + \bw_t,
\label{global_update_}
\end{equation}
where $\gm(.)$ stands for the geometric median aggregation defined as
$$
\gm(\{ \bu_i\}_{i\in \mathcal{K}}) = {\rm arg} \min_\bz \sum_{i\in \mathcal{K}}  \|\bz-\bu_i\|.
$$
The geometric median  aggregation has been proposed as an efficient solution to Byzantine attacks when the individual updates of the clients are sent separately to the CS. In fact, it approximates well the mean of the honest clients weight updates when $B<N/2$ \cite{9153949}. In our case, we will use it to aggregate the group updates. The number of Byzantine workers should satisfy $B<G/2$ in order for the geometric median to well approximate the mean of the group updates composed by only honest clients.

To compute the geometric median, the Weiszfeld algorithm can be used \cite{weiszfeld1937point}. To avoid numerical instabilities, a smoothed version of the Weiszfeld algorithm can be implemented in practice \cite{pillutla2019robust}, which computes a smoothed geometric median defined as 
$
\gm_{\epsilon}(\{ \bu_i\}_{i\in \mathcal{K}}) = {\rm arg} \min_\bz \sum_{i\in \mathcal{K}}  \|\bz-\bu_i\|_\epsilon,
$
where
$$
\|\bx\|_\epsilon=\begin{cases}
\frac{1}{2\epsilon}\|\bx\|^2+\frac{\epsilon}{2} \ \ \ {\rm if} \ \ \|\bx\|\leq\epsilon\\ 
\|\bx\| \ \ \ \ \ \ \ \ \ \ \ \ \    {\rm if} \ \ \|\bx\|>\epsilon,
\end{cases}
$$
where $\epsilon>0$ is a smoothing parameter.
The steps of the proposed approach are summarised in Algorithm \ref{sum_alg}, where $LocalComp(\bw_t,H,b,\eta,D_n)$ consists of $H$ steps  of batch-SGD using local dataset $D_n$ with learning rate $\eta$ and initial parameter vector $\bw_t$ as described in \eqref{SGD}.

\begin{algorithm}
  \SetAlgoLined
  \caption{}
  \BlankLine
\KwIn{Initial model $\bw_0$}
\For{$t = 0 , 1, 2 ,\cdots$}{The CS forwards $\bw_t$ to the clients\;
\For{each client $n$}{
$\bm_t^n\gets LocalComp(\bw_t,H, b,\eta,D_n)$
}
\For{$g = 1,2,\cdots,G$}{
	\For{each client $n $ in group $g$ $(n\in \mathcal{G}_{g,t}) $}{
	client $n$ transmit its model update $\bx_{t,n}$ precoded via \eqref{precoding_11} during transmission time slot $T_g$
	}
	The CS receives $\by_g$ of group $g$ and computes $\bu_t^g$ as in \eqref{group_update}
}
The CS aggregates the group updates using \eqref{global_update_} to obtain the new global parameter vector $\bw_{t+1}$
}
\label{sum_alg}
\end{algorithm}

\section{Convergence Analysis}
\label{conver_analysis}
In this section, we analyze the convergence of the proposed approach. Several assumptions on the loss functions are needed.
\begin{assumption}
\begin{itemize}
\item[(i)] Strong convexity: The objective function $f$ is $\mu$-strongly convex, that is, for all $\bx,\by \in \mathbb{R}^p$
$$
f(\bx)\geq f(\by) + \langle f'(\by), \bx-\by\rangle+\frac{\mu}{2}\|\bx-\by\|^2.
$$
for some $\mu>0$.
\item[(ii)] Lipschitz continuity of gradients: The objective function $f$ has $L$-Lipchitz continuous gradients, that is, for all $  \bx,\by \in \mathbb{R}^p$
$$
\|f'(\bx)-f'(\by)\| \leq L \|\bx-\by\|.
$$
for some $L>0$.
\item[(iii)] Bounded outer variation: For each honest client $n$, the variations of its aggregated gradients with respect to the over-all gradient is bounded as
$$
\mathbb{E}\| f'_{n}(\bx)-f'(\bx) \|^2 \leq \delta^2,   \  {\rm for   \ all} \  \bx \in \mathbb{R}^p.
$$
\item[(iv)] Bounded inner variation: for each honest client $n$, the variation of its stochastic gradients with respect to its aggregated gradients is bounded as
$$
\mathbb{E}\|f'_{n,i_n^t}(\bx)-f_n'(\bx)\|^2\leq \kappa^2,  \  {\rm for   \ all} \  \bx \in \mathbb{R}^p.
$$
\item[(v)] Bounded stochastic gradients: For each honest client $n$, stochastic gradient $f'_{n,i_n^t}(\bx)$ satisfies
$$
\mathbb{E}\|f'_{n,i_n^t}(\bx)\|^2\leq K^2, \  {\rm for   \ all} \  \bx \in \mathbb{R}^p,
$$
for some fixed $K^2>0$.
\end{itemize}
\label{assump1}
\end{assumption}
Items (i) and (ii) in Assumption \ref{assump1} are common in convex analysis. Items (iii) and (iv) are needed to bound the inner and outer variations of the stochastic gradients and the gradients of the honest clients, respectively. These assumptions are adopted in most of the existing works considering distributed SGD in presence of Byzantine attacks \cite{9153949,pmlr-v80-tang18a}. The convergence of the proposed approach is presented in the following theorem. For simplicity, we assume in this section that the learning rate is constant and that the clients perform one SGD step at each global iteration, that is, $H=1$ and $\eta_t=\eta$ for all $t$.
\begin{theorem} Under Assumption \ref{assump1}, when the number of Byzantine attackers satisfies $B< \frac{G}{2}$ and the step size $\eta$ verifies $\eta< \frac{\mu}{2L^2}$, then 
$$
\mathbb{E}\|\bw_t-\bw^*\|^2\leq  ( 1-\eta\mu)^{t} B+A,
$$
where 
\begin{align*}
B&= \|\bw_0-\bw^*\|^2 -A,\\
A &= \frac{2}{\mu^2}C_\alpha^2 \left(\delta^2+\kappa^2+\frac{p\sigma^2}{Ph_{min}^2} K^2\right),
\end{align*}
with $C_\alpha=\frac{2-2\alpha}{1-2\alpha}$ and $\alpha= \frac{B}{G}$.
\label{thm1}
\end{theorem}
\begin{IEEEproof}
See Appendix.
\end{IEEEproof}
Theorem \ref{thm1} states that the proposed approach converges at a linear rate to a neighborhood of $\bw^*$. The asymptotic learning error, $A$, depends on the number of Byzantine attackers through $C_\alpha$ and on the noise variance. When the number of Byzantine attackers increases, $C_\alpha$ increases, which yields a higher asymptotic error.

\section{Numerical Results}
\label{numerical_simulation}
In order to evaluate the performance of the proposed approach, we provide in this section simulation on real dataset. We consider the MNIST dataset composed of $28 \times 28$ images of handwritten digits corresponding to 10 classes (digits 0 to 9). The dataset set contains $60, 000$ training samples and 10,000 testing samples. We divided the training samples equally at random to $K=100$ clients, that is, each client has a local dataset composed of $600$ samples. We used the multi-class logistic regression model. At every global training round, each client performs $H = 12$ steps of local batch-SGD where at each step a minibatch of size $b=50$ is used. The leaning rate $\eta_t$ is fixed to $0.01$ in all local SGD steps.The number of groups is fixed to $G=20$, while the noise variance and the power constraint are set to be $\sigma^2=10^{-2}$ and $P=1$, respectively. The smoothing parameter of the Weiszfeld algorithm used is $\epsilon=10^{-4}$. In all simulations, the transmission threshold $h_{\min}$ and the scaling factor $\rho_t$ are fixed to 0.1 and 8 respectively.

In the first experiment, we consider Gaussian attacks where each Byzantine client sends a Gaussian vector of mean $\bw_t$ and variance $30$ instead of its actual model update. We compare our proposed approach and the simple averaging COTAF \cite{Sery2020} described in Section \ref{system_model}. From \figref{gaussian_attacks}, simple averaging is not robust to Byzantine attacks. On the other hand, our proposed approach guarantees fast convergence.

In the second experiment, we consider class flip attacks where the Byzantine clients change the labels of their local datasets as $y=9-y$.  \figref{class_flip_attacks} demonstrates the effect of increasing the number of Byzantine workers. From this figure, the proposed approach is robust to Byzantine attacks even though the test accuracy relatively decreases with the number of the Byzantine clients. This is expected since the training, in presence of malicious clients, is done over a smaller sample size as only the honest workers contribute in the learning process. This was also predicted by the convergence analysis conducted in the previous section, where it has been shown that the asymptotic learning error increases with the number of Byzantine attacks.

\begin{figure}
\begin{center}
\subfigure[Test Loss]{
}
\end{center}
\caption{Test loss and test accuracy vs. the number of iterations when $B=9$ Byzantine clients applying Gaussian attacks.}
\label{gaussian_attacks}
\end{figure}

\begin{figure}[ht]
\begin{center}
\subfigure[Test Loss]{
%
}
\end{center}
\caption{Test loss and test accuracy vs. the number of iterations for different numbers of Byzantine clients applying class-flip attacks.}
\label{class_flip_attacks}
\end{figure}

\section{Conclusion}
\label{conclusion}
In this paper, we have proposed a novel approach to account for Byzantine attacks in over-the-air FL. By dividing the distributed clients into groups, the parameter server is able to reduce the effect of Byzantine attacks by aggregating the group parameter updates. The convergence of the proposed algorithm has been studied analytically under assumption of convex loss functions. The simulation results show the robustness of the approach to different Byzantine attacks. This work can be extended by studying other aggregation techniques. It is also important to study the effect of the imperfect channel knowledge on the proposed algorithm.

\section*{Appendix}
We first sate the following lemma which will be used later in the proof.
\begin{lemma}\cite[Lemma 2]{9153949} Let $\mathcal V$ be a subset of random vectors distributed in
a normed vector space. If $\mathcal{V}' \subset \mathcal{V}$ such that $|\mathcal{V}'| < \frac{|\mathcal{V}|}{2}$,
then it holds that
$$
\mathbb{E}\|\underset{\bv\in \mathcal{V}}\gm (\bv) \|^2 \leq C_\alpha^2 \frac{\sum_{\bv \notin \mathcal{V}'}\|\bv\|^2}{|\mathcal{V}| - |\mathcal{V}'| },
$$
where $C_\alpha=\frac{2-2\alpha}{1-2\alpha}$ and $\alpha= \frac{|\mathcal{V}'| }{|\mathcal{V}| }$.\label{gm_Byzantine}\end{lemma}
Define $\delta_t \triangleq \mathbb{E}\|\bw_t-\bw^*\|^2$. To prove Theorem 1, we start by finding an upper bound for $\delta_{t+1}$,
\begin{align*}
\delta_{t+1} & = \mathbb{E}\|\bw_{t+1}-\bw^*\|^2\\&= \mathbb{E}\|\bw_t-\eta f'(\bw_t)-\bw^*+\bw_{t+1}-\bw_t+\eta f'(\bw_t)\|^2\\
&\leq \frac{1}{1-\gamma}\mathbb{E}\|\bw_t-\eta f'(\bw_t)-\bw^*\|^2\\& \ \ \ \ +\frac{1}{\gamma}   \mathbb{E}\|\bw_{t+1}-\bw_t+\eta f'(\bw_t)\|^2,
\end{align*}
for any $0<\gamma<1$, where we have used the fact that $\|\bx+\by\|^2\leq \frac{1}{1-\gamma}\|\bx\|^2+\frac{1}{\gamma}\|\by\|^2$ for any $0<\gamma<1$. Since $f'(\bw^*)=0$, we can write
\begin{align*}
&\|\bw_t-\eta f'(\bw_t)-\bw^*\|^2 \\&=\|\bw_t-\eta ( f'(\bw_t)-f'(\bw^*) )-\bw^*\|^2\\ &= 
\|\bw_t-\bw^*\|^2 -2 \eta \langle f'(\bw_t)-f'(\bw^*), \bw_t -\bw^*\rangle \\&  \ \ \ \ \ + \eta^2 \|f'(\bw_t)-f'(\bw^*)\|^2
\\ & \overset{(a)}{\leq}  \|\bw_t-\bw^*\|^2 -2 \eta \mu  \|\bw_t-\bw^*\|^2 + \eta^2 L^2 \|\bw_t-\bw^*\|^2
\end{align*}  
where $(a)$ follows from items $(i)$ and $(ii)$ of Assumption \ref{assump1}. Thus,
\begin{align*}
&\delta_{t+1} \leq \frac{1-2\eta\mu +\eta^2L^2}{1-\gamma}\delta_t + \frac{1}{\gamma}   \mathbb{E}\|\bw_{t+1}-\bw_t+\eta f'(\bw_t)\|^2.
\end{align*}  
For  $\eta<\frac{2}{\mu}$, we can take $\gamma = \frac{\eta\mu}{2}$. Assuming further that $\eta \leq \frac{\mu}{2L^2}$, it holds that
$
\frac{1-2\eta\mu +\eta^2L^2}{1-\gamma} \leq 1-\eta\mu
$. Hence, for $\eta <\min(\frac{2}{\mu}, \frac{\mu}{2L^2}) = \frac{\mu}{2L^2}$,
\begin{align}
\delta_{t+1}  &\leq ( 1-\eta\mu) \delta_t   +\frac{2}{\eta \mu}   \mathbb{E}\|\bw_{t+1}-\bw_t+\eta f'(\bw_t)\|^2.
\label{eq:102}
\end{align}
We treat now the second term of the right-hand side of \eqref{eq:102}. From the update rule \eqref{global_update_}, it follows that
$
\bw_{t+1} -\bw_{t} = \gm(\{\bu_g^t\}_{g=1}^G)  
$
 where $\bu_g^t$, for the case $H=1$, is given by
 \begin{align*}
 \bu_g ^t&=  -\frac{1}{|\mathcal{K}_{t,g}|} \sum_{n\in \mathcal{K}_{t,g}} \eta f_{n,i_n^t}'(\bw_t) + \bz_{t,g},
\end{align*}
Define $\mathcal{B}_t$ as the set of groups containing at least one Byzantine attacker at global iteration $t$.
Applying Lemma \ref{gm_Byzantine} yields
{\small
\begin{align*}
& \mathbb{E}\|\bw_{t+1}-\bw_t+\eta f'(\bw_t)\|^2  =  \mathbb{E}\|\gm(\{\bu_g^t\}_{g=1}^G)+\eta f'(\bw_t)\|^2\\&
 =  \mathbb{E}\|\gm(\{\bu_g^t+\eta f'(\bw_t)\}_{g=1}^G)\|^2\\&
  \leq C_\alpha^2 \frac{\sum_{g \notin \mathcal B_t}\mathbb{E}\left\| \bu_{g}^t + \eta f'(\bw_t) \right\|^2}{G-|\mathcal B_t|},
 \end{align*}}
Replacing $\bu_{g}^t$ by its expression, it follows that 
 {\small
  \begin{align*}
  &\mathbb{E}\|\bw_{t+1}-\bw_t+\eta f'(\bw_t)\|^2 \\&\leq C_\alpha^2 \frac{\sum_{g \notin \mathcal B_t}\mathbb{E} \left\|-\frac{\eta}{|\mathcal K_{t,g}|} \sum_{n \in  \mathcal K_{t,g}}  f'_{i^n}(\bw_t) + \bz_{t,g} + \eta f'(\bw_t) \right\|^2}{G-|\mathcal B_t|} \\&
 \leq  C_\alpha^2 \eta^2 \frac{\sum_{g \notin \mathcal B_t} \frac{1}{|\mathcal K_{t,g}|} \sum_{n \in  \mathcal K_{t,g}} \mathbb{E}\left\| f'_{n,i_n^t}(\bw_t)- f'_n(\bw_t) \right\|^2}{G-|\mathcal B_t|}
 \\& +  C_\alpha^2 \eta^2 \frac{\sum_{g \notin \mathcal B_t} \frac{1}{|\mathcal K_{t,g}|} \sum_{n \in  \mathcal K_{t,g}}\mathbb{E}\left\| f'_{n}(\bw_t) - f'(\bw_t) \right\|^2}{G-|\mathcal B_t|}
 \\&+ C_\alpha^2  \frac{\sum_{g \notin \mathcal B_t}\mathbb{E}\|\bz_{t,g}\|^2}{G-|\mathcal B_t|},
\end{align*}}
where the last result follows from the fact that $\left\|\sum_{i=1}^k \bv_i \right\|^2\leq k \sum_{i=1}^k\|\bv_i\|^2$ for any sequence of vectors $\{\bv_i\}_{i=1}^k$ and the independence between the noise vectors and the stochastic gradients. $\mathbb{E}\|\bz_{t,g}\|^2$ can be bounded as follows
\begin{align}
&\mathbb{E}\|\bz_{t,g}\|^2= \frac{p\sigma^2}{h_{min}^2 |\mathcal K_{t,g}|^2\rho_t^2} =  \frac{p\sigma^2}{h_{min}^2|\mathcal K_{t,g}|^2} \frac{\max_n\mathbb{E}\|\bm_n^t\|^2}{P}\nonumber
\\& \leq \frac{p\sigma^2}{Ph_{min}^2|\mathcal K_{t,g}|^2}\eta^2 K^2 \leq \frac{p\sigma^2}{Ph_{min}^2}\eta^2 K^2.
\label{noise_bound}
\end{align}
Combining \eqref{noise_bound} and items (iii) and (iv) in Assumption \ref{assump1}, it holds that
{\small
\begin{align}
 \mathbb{E}\|\bw_{t+1}-\bw_t+\eta f'(\bw_t)\|^2 \!\leq C_\alpha^2 \eta^2\! \left(\delta^2+\kappa^2+\frac{p\sigma^2}{Ph_{min}^2} K^2\right)\!.
 \label{eq202}
\end{align}}
Combining \eqref{eq:102} and \eqref{eq202} yields for $\eta<\frac{\mu}{2L^2}$
$$
\delta_{t+1}\leq  ( 1-\eta\mu)   \delta_t +\eta \mu A,
$$
where
$
A  \triangleq \frac{2}{\mu^2}C_\alpha^2 \left(\delta^2+\kappa^2+\frac{p\sigma^2}{Ph_{min}^2} K^2\right)
$. Thus,
\begin{align}
\delta_{t+1} & \leq ( 1-\eta\mu)^{t+1} \delta_0 + \eta \mu A\sum_{i=0}^{t}( 1-\eta\mu)^{i}\\
& =  ( 1-\eta\mu)^{t+1} ( \delta_0 - A)+A,
\end{align}
which completes the proof.

\bibliographystyle{IEEEtran}
\bibliography{IEEEabrv,IEEEconf,../references}
\end{document}